\documentclass[letterpaper, 10 pt, conference]{ieeeconf}  

\IEEEoverridecommandlockouts                              
\usepackage{graphics} 
\usepackage{epsfig} 
\usepackage{mathptmx} 
\usepackage{times} 
\usepackage{amsmath} 
\usepackage{amssymb}  
\usepackage{amsfonts}
\usepackage{algorithmic}
\usepackage{graphicx}
\usepackage{textcomp}
\usepackage{xcolor}
\def\BibTeX{{\rm B\kern-.05em{\sc i\kern-.025em b}\kern-.08em
    T\kern-.1667em\lower.7ex\hbox{E}\kern-.125emX}}

\usepackage{color}
\usepackage{subfig}
\usepackage{subfloat}
\usepackage{multirow} 
\usepackage{arydshln}

\graphicspath{{figures/}}

\title{\LARGE \bf
A Bearing-Strength Method for Motion Estimation \\of Unknown Energy Emitters
}

\author{Haoyu Chen, Zian Ning, Yin Zhang, Shiyu Zhao$^\dagger$
\thanks{$^\dagger$ Corresponding author}%
\thanks{All the authors are with the WINDY Lab in the School of Engineering at Westlake University, Hangzhou, China. S. Zhao is also with the Research Center for Industries of the Future, Westlake University, Hangzhou, China.
\{ningzian, chenhaoyu, zhangyin, zhaoshiyu\}@westlake.edu.cn}}

\begin{document}

\maketitle
\thispagestyle{empty}
\pagestyle{empty}

\begin{abstract}
This paper studies motion estimation of moving energy emitters using passive sensors. The emitters may be light, acoustic, or radio sources. While the bearing vector pointing from the sensor to the emitter can be easily obtained, existing approaches mainly rely on the bearing-only motion estimation method. However, this method suffers from a fundamental limitation that the sensor must have lateral motion to ensure observability. Unfortunately, this lateral motion requirement often conflicts with the sensor's desired motion in many tasks. In this paper, we point out that the received signal strength, which can also be obtained easily in many ways, can greatly enhance motion estimation. Surprisingly, this strength information has not been well explored so far. Here, we propose a new bearing-strength method to fully exploit both the bearing and strength measurements. Our theoretical analysis shows that the system observability is significantly enhanced in the sense that the lateral motion condition is not required anymore. Real-world experimental results verify the proposed method and the theoretical analysis. It is notable that the benefit of the proposed method comes with no additional cost since it simply utilizes the received strength information that has not been fully exploited in the past.
\end{abstract}

\section{INTRODUCTION}

This paper studies the problem of estimating the motion of a moving emitter using a moving passive sensor.
The emitter may be a light, acoustic, or radio source, whose true transmit power is \emph{unknown} in advance.
The specific motivation for our study is aerial target pursuit, where one flying MAV (micro aerial vehicle) uses its onboard device to detect, localize, and capture another flying target MAV~\cite{10659110, Ning2024a}.
Our recent works proposed motion estimation methods based on RGB vision detection of MAVs~\cite{Ning2024}. 
However, it is nontrivial to obtain highly reliable RGB vision detection results in complex scenarios~\cite{Vrba2020, guo2024global, zhang2024domain}. 
We therefore have been exploring other types of measurements, such as signal strength for target motion estimation. 
Beyond aerial target pursuit, the study in this paper is also applicable to other tasks since emitter localization is a fundamental problem that widely exists in many applications~\cite{Hoelzer1978, zanella2016best}. 

\begin{figure}[!t]
    \centering
    \includegraphics[width=\linewidth]{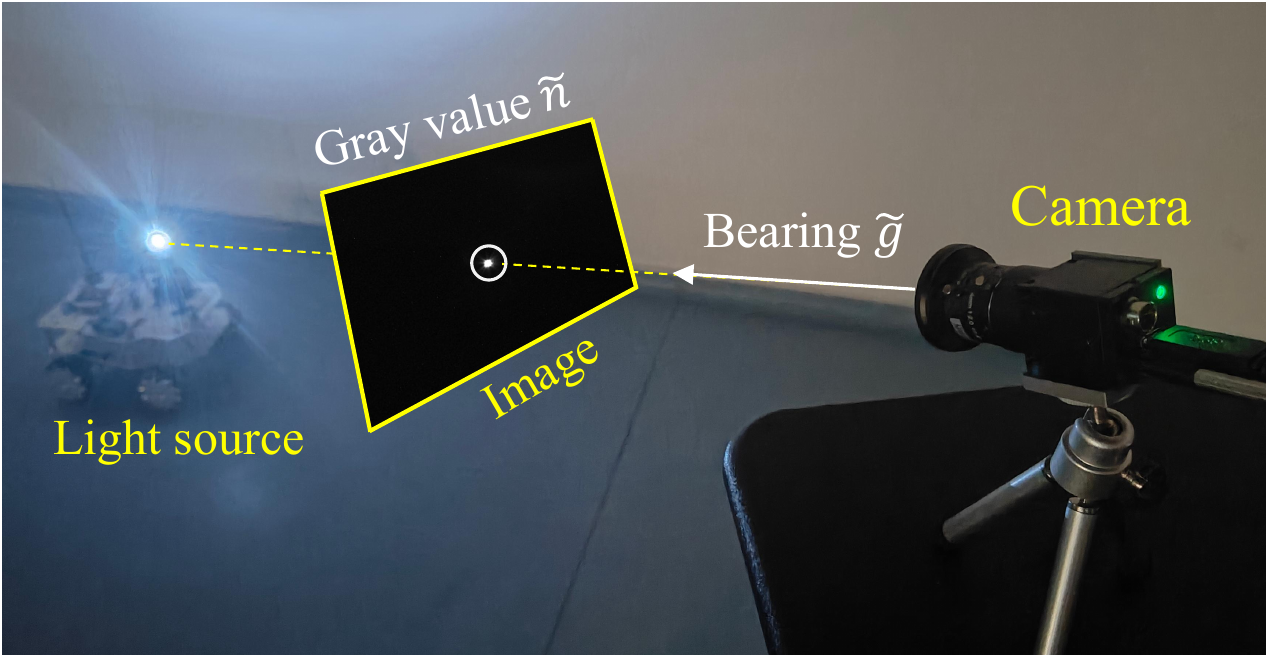}
    \caption{A moving camera observes a moving light source. The bearing $\tilde{g}$ and gray value $\tilde{n}$ can be obtained based on the brightest optical flare in the image.}
    \label{fig_indoor}
\end{figure}

A passive sensor usually can obtain two types of valuable information about emitters (see Fig.~\ref{fig_indoor}). 

The first is the bearing vector pointing from the sensor to the emitter. 
For acoustic and radio sources, the bearing vector can be obtained based on sensor arrays ~\cite{Dressel2019, Shi2018}. 
For light sources, the bearing vector can be obtained from the pixel coordinate of the emitter's image~\cite{Ning2024}.
Using the bearing vector to estimate the emitter's motion is known as the bearing-only method, which was initially developed for estimating the motion of ships~\cite{Hoelzer1978} and regained attention in recent years in vision-based motion estimation~\cite{He2019, Li2023, Ning2024}.
A limitation of the traditional bearing-only method is that the sensor must move in the lateral direction that is orthogonal to the bearing vector~\cite{Fogel1988}.
Such lateral motion usually conflicts with the desired motion of the sensor in many tasks~\cite{Li2023}.
It is therefore important to study other ways that can ensure observability while avoiding unfavorable lateral motion.

Our recent work~\cite{Ning2024} explored a bearing-angle approach to address the observability deficiency in radial directions by utilizing the target's apparent orientation. While effective, the bearing-angle method~\cite{Ning2024} necessitates sophisticated image recognition to extract the target's geometric features, which imposes a significant computational burden on resource-constrained MAV platforms. Furthermore, the reliability of angle extraction degrades substantially in low-light environments or complex backgrounds where target silhouettes are blurred.

\begin{figure}[!t]
    \centering
    \includegraphics[width=0.9\linewidth]{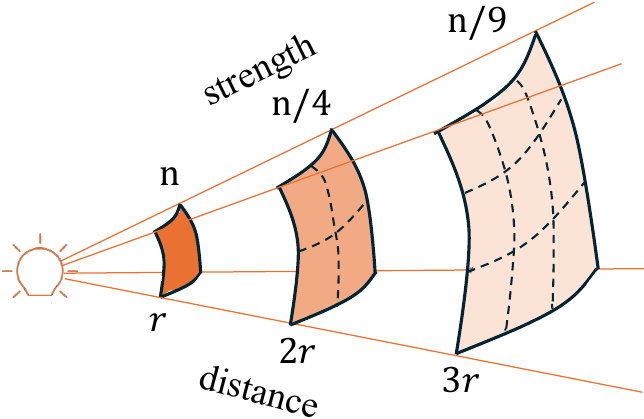}
    \caption{The strength of point-like energy emitters measured by passive sensors is inversely proportional to the square of their distance.}
    \label{fig_square_inverse}
\end{figure}

The second type of information is the received signal strength (RSS).
For a point-like emitter, the received strength is inversely proportional to the square of the distance to the emitter~\cite{Voudoukis_Oikonomidis_2017} (see Fig.~\ref{fig_square_inverse}).
As a result, the received strength is jointly determined by the emitter's distance and transmit power.
The RSS measurement has been widely used in node localization problems within wireless sensor networks, where multiple sensor devices are used simultaneously~\cite{zanella2016best, wang2011new, tomic20163}.
However, they cannot handle the case where there is only one passive sensor.
Also, surprisingly, the RSS-like measurement has not been adopted to estimate the motion of light sources up to now, to the best of our knowledge.

This paper proposes a novel bearing-strength method that exploits both bearing and received strength measurements. Compared to our recent bearing-angle approach~\cite{Ning2024}, which enhances observability through geometric features, the proposed method offers several distinct advantages. First, by relying on pixel intensity (RSS) rather than complex geometry, it significantly reduces the computational overhead, making it more suitable for resource-constrained MAVs. Second, detecting a light spot for strength measurement is inherently more robust in low-light conditions, where high contrast facilitates detection even when the target's silhouette is indiscernible. Finally, we theoretically demonstrate that the bearing-strength method provides a direct physical constraint on the radial distance, ensuring observability without the need for lateral motion.

The technical novelties of this framework are detailed as follows:

First, the strength measurement does not directly infer the emitter's distance, as the original emitter's power is \emph{unknown}. 
It is therefore nontrivial to utilize the received strength measurement. 
The key here is to convert the nonlinear strength measurement to be pseudo-linear. Moreover, we augment the state vector that consists of the emitter's position and velocity by a new strength variable corresponding to the unknown transmit power of the emitter. Then, pseudo-linear Kalman filtering is applied to fuse the measurements and estimate the emitter's motion.

Second, although a new strength measurement is introduced to the estimator, an additional \emph{unknown} about the emitter's power should also be estimated. It is therefore nontrivial to see how the additional strength measurement can help improve the system's observability.
We conduct a theoretical analysis and show that the observability is ensured if and only if the sensor has higher-order motion than the emitter. This condition does not require the sensor to move in the lateral direction that is orthogonal to the bearing vector, significantly weakening the observability condition.

Third, our method is general in the sense that it can apply to various energy emitters such as acoustic, radio, and light sources. However, this paper considers a special yet important energy emitter: light sources. We show that a grayscale camera can be used as a luminance meter to sense the strength of the light source. Our experiments show that the gray value of the light source's image is proportional to the received light strength~\cite{wuller2007}. Moreover, the detection of the optical flare in a grayscale image is usually much easier than detecting a complex object in an RGB image.

Real-world experiments have verified the effectiveness of the bearing-strength method and the theoretical observability analysis. It is notable that the benefit of the proposed method comes with no additional cost since it simply exploits the received strength information that has not been fully exploited in the past.

\section{Problem Statement}

Energy emitters exhibit different radiation patterns depending on their physical nature. Acoustic and radio sources are often treated as isotropic point sources in many applications. In contrast, light sources typically exhibit anisotropy. However, our proposed bearing-strength framework is designed to be general enough to handle both cases.

We consider a special yet important type of emitters, \emph{light sources}, in the rest of the paper, although the proposed method is also applicable to other types of emitters.
Consider a light source that moves in 3D space. We suppose the light source is a point mass. Even if the light source occupies a certain volume, it can still be approximated as a point mass when observed from a distance.
Denote the emitter's position and velocity as $p_e\in\mathbb{R}^3$ and $v_e\in\mathbb{R}^3$, respectively.
They are the two unknowns that we aim to estimate.

The passive sensor is a monocular grayscale camera. 
When a camera observes a light source, an optical flare appears in the image.
The optical flare can be detected in the image in various ways. 
For example, we can select the brightest pixel.
In general, it is easier to detect the optimal flare than to detect a complex object in an RGB image~\cite{guo2024global, zhang2024domain}. 
Notably, the use of optical flares offers the advantage of handling low-light conditions.
Optical filters may be used under challenging lighting conditions, which is not in the scope of this paper.
Moreover, the sensor's pose is assumed to be available. This is a reasonable assumption in many applications where it can be either obtained via RTK GPS~\cite{Li2023, Ning2024} or estimated by visual inertial odometry~\cite{Qiu2019}.

Two types of useful information can be extracted from the optical flare in the image.
The first information is the bearing measurement. Let the unit vector $\tilde{g}\in\mathbb{R}^3$ represent the bearing vector pointing from the camera to the emitter. It can be calculated based on the pixel coordinate of the optical flare in the image $q_\mathrm{pix}=[x_\mathrm{pix}, y_\mathrm{pix}, 1]^\mathrm{T}\in\mathbb{R}^3$, the intrinsic parameter matrix of the camera $P_\mathrm{cam}$, and the attitude of the camera $R_\mathrm{c}^\mathrm{w}\in\mathbb{R}^3$ \cite{Ning2024}:
\begin{align*}
\tilde{g}=\dfrac{R_\mathrm{c}^\mathrm{w}P_\mathrm{cam}^{-1}q_\mathrm{pix}}{\|R_\mathrm{c}^\mathrm{w}P_\mathrm{cam}^{-1}q_\mathrm{pix}\|}\in\mathbb{R}^3.
\end{align*}
The second piece of information is the received strength measurement. 
In particular, let 
\begin{align*}
    \tilde{n}>0
\end{align*}
be the gray value of the pixel of the optical flare in the image. We will show later that $\tilde{n}$ is proportional to the received light strength.

Our goal is to estimate the emitter's position $p_e$ and velocity $v_e$ based on the noisy bearing vector $\tilde{g}$, and the strength measurement $\tilde{n}$, together with the sensor's position $p_s$.

\section{Bearing-Strength Motion Estimator}

This section presents the new bearing-strength motion estimator. The key is to establish appropriate measurement equations and state transition equations.

\begin{figure*}
    \centering
    \includegraphics[width=\linewidth]{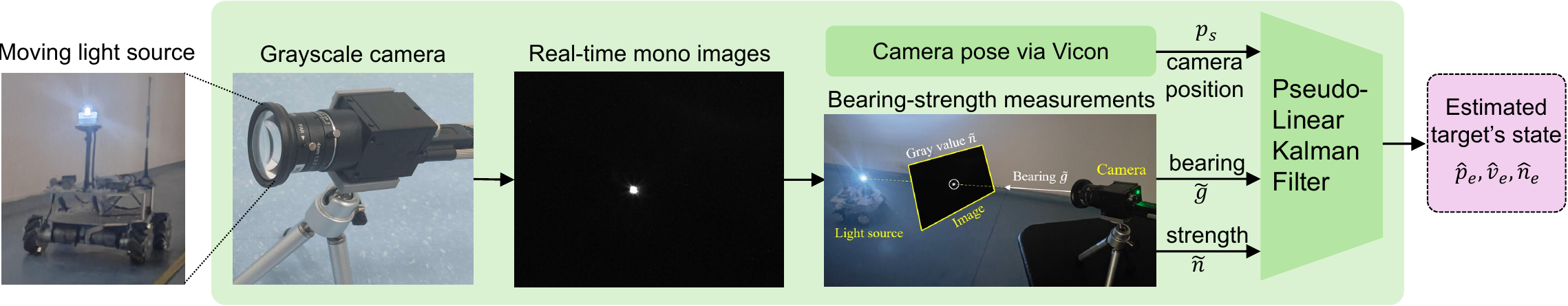}
    \caption{The architecture of the proposed bearing-strength method.}
    \label{fig_algorithm}
\end{figure*}

\subsection{State transition equation}

The emitter's state vector is designed as
\begin{align*}
    x=\begin{bmatrix}
        p_e^\mathrm{T},v_e^\mathrm{T},\sqrt{n_e}
    \end{bmatrix}^\mathrm{T}\in\mathbb{R}^7.
\end{align*}
Here $n_e>0$ is an \emph{unknown} scalar corresponding to the emitter's power.
Specifically, it equals the gray value of the optical flare in the image when the camera is one meter away from the light source. 
The square root for $n_e$ is introduced to facilitate the subsequent pseudo-linearization of the nonlinear measurement equations. 

It is common to model the emitter's motion as a noise-driven discrete-time double integrator~\cite{Li2023, Ning2024}:
\begin{align}
    \label{eq_state_transition}
    x(t_{k+1})=Fx(t_k)+Bq(t_k),
\end{align}
where
\begin{align*}
    F&=\begin{bmatrix}
        I_{3\times3} & \delta tI_{3\times3} & 0_{3\times 1} \\
        0_{3\times3} & I_{3\times3} & 0_{3\times 1} \\
        0_{1\times3} &0_{1\times3} &1
    \end{bmatrix}\in\mathbb{R}^{7\times7},\\ 
    B&=\begin{bmatrix}
        \dfrac{1}{2}\delta t^2I_{3\times3} & 0_{3\times 1} \\
        \delta tI_{3\times 3} & 0_{3\times 1}  \\
        0_{1\times 3} & 1
    \end{bmatrix}\in\mathbb{R}^{7\times 4}.
\end{align*}
Here, $\delta t$ is the sampling time, and $I$ and $0$ are identity and zero matrices, respectively.
Moreover, $q\sim\mathcal{N}(0, \Sigma_q)\in\mathbb{R}^4$ is the process noise, whose covariance matrix is $\Sigma_q = \text{diag}(\sigma_a^2, \sigma_a^2,\sigma_a^2,\sigma_{n}^2)\in\mathbb{R}^{4\times 4}$. 

\subsection{Nonlinear measurement equations}

The bearing vector $\tilde{g}$ and the gray value $\tilde{n}$ are both nonlinear functions of the emitter's position. In particular,
\begin{align}
    \tilde{g}&=\dfrac{p_e-p_s}{r}+\mu_g,\label{eq_g_nonlinear_mear_eq}\\
    \sqrt{\tilde{n}}&=\dfrac{\sqrt{n_e}}{r}+\mu_n, \label{eq_strength_nonlinear_mear_eq}
\end{align}
where $r=\|p_e-p_s\|\in\mathbb{R}$ is the distance between the emitter and the sensor. Here, $\mu_g\sim\mathcal{N}(0,\sigma^2_gI_{3\times3})\in\mathbb{R}^3$ and $\mu_n\sim\mathcal{N}(0, \sigma^2_n)\in\mathbb{R}$ are measurement noises.
Noted that the additive noise $\mu_g$ in~\eqref{eq_strength_nonlinear_mear_eq} is obtained by transforming the original productive noise~\cite{Li2023} and can be approximately treated as a Gaussian noise~\cite{Li2023, Ning2024}. 

Some remarks about \eqref{eq_strength_nonlinear_mear_eq} are given below.
First, the camera parameters must be fixed all the time so that the gray values are proportional to the measured light strength~\cite{hiscocks2011measuring, wuller2007}.
Second, the light strength obeys the inverse distance square law~\cite{Voudoukis_Oikonomidis_2017} (see Fig.~\ref{fig_square_inverse}). 
Detailed tests are shown in Fig.~\ref{fig_experiment_scenario}.

\subsection{Pseudo-linear measurement equations}

First, we convert the nonlinear bearing measurement in \eqref{eq_g_nonlinear_mear_eq} to be pseudo-linear. To do that, multiplying $r(I-\tilde{g}\tilde{g}^\mathrm{T})$ on both sides of~\eqref{eq_g_nonlinear_mear_eq} yields
\begin{align}
    \label{eq_g_pseudo}
    P_{\tilde{g}}p_s=P_{\tilde{g}}p_e + rP_{\tilde{g}}\mu_g,
\end{align}
where $P_{\tilde{g}}=I-\tilde{g}\tilde{g}^\mathrm{T}\in\mathbb{R}^{3\times3}$ is an orthogonal projection matrix, which plays an important role in the bearing-based estimation problems~\cite{Zhao2019}.
Detailed pseudo-linear transformation of the bearing measurement in \eqref{eq_g_nonlinear_mear_eq} is omitted due to space limitations and can be found in our previous works~\cite{Li2023, Ning2023, Ning2024}.

Second, we convert the nonlinear strength measurement in~\eqref{eq_strength_nonlinear_mear_eq} to be pseudo-linear.
Multiplying $r\tilde{g}$ on both sides of~\eqref{eq_strength_nonlinear_mear_eq} yields
\begin{align}
    \label{eq_nd_tem2}
    r\tilde{g}\sqrt{\tilde{n}}=\tilde{g}\sqrt{n_e}+r\tilde{g}\mu_n.
\end{align}
It follows from~\eqref{eq_g_nonlinear_mear_eq} that $r\tilde{g}=p_e-p_s+r\mu_g$, substituting which into the left side of~\eqref{eq_nd_tem2} and reorganizing it gives
\begin{align}
    \label{eq_nd_pseudo}
    \sqrt{\tilde{n}}p_s=\sqrt{\tilde{n}}p_e-\tilde{g}\sqrt{n_e} + r\left(\sqrt{\tilde{n}}\mu_g-\mu_n\tilde{g}\right).
\end{align}
Equations \eqref{eq_g_pseudo} and \eqref{eq_nd_pseudo} are called ``pseudo-linear'' because the measurements also appear on the right-hand side of equations.

Combining~\eqref{eq_g_pseudo} and~\eqref{eq_nd_pseudo} gives a compact matrix-vector form of the pseudo-linear measurement equation:
\begin{align}
    \label{eq_final_measurement_eq}
    z=Hx+\nu,
\end{align}
where
\begin{align*}
    z&=\begin{bmatrix}
        P_{\tilde{g}}p_s \\
        \sqrt{\tilde{n}}p_s
    \end{bmatrix}\in\mathbb{R}^6, \\
    H&=\begin{bmatrix}
        P_{\tilde{g}} & 0_{3\times3} & 0_{3\times 1}\\
        \sqrt{\tilde{n}}I_{3\times3} & 0_{3\times3} & -\tilde{g}
    \end{bmatrix}\in\mathbb{R}^{6\times 7}, \\
    \nu&=r\begin{bmatrix}
    P_{\tilde{g}}\mu_g \\
    \sqrt{\tilde{n}}\mu_g-\mu_n\tilde{g}
    \end{bmatrix}=E
    \begin{bmatrix}
        \mu_g \\ \mu_n
    \end{bmatrix}\in\mathbb{R}^{6}.
\end{align*}
Here, $\nu$ can be treated as a linear transformation of Gaussian noises with
\begin{align*}
    E = r\begin{bmatrix}
        P_{\tilde{g}}\mu_g & 0_{3\times 1}\\
         \sqrt{\tilde{n}}I_{3\times 3} & -\tilde{g}
    \end{bmatrix}\in\mathbb{R}^{6\times 4}.
\end{align*}
The covariance matrix of $\nu$ can be calculated as
\begin{align*}
    \Sigma_\nu = E
    \begin{bmatrix}
        \sigma_g^2 I_{3\times3} & 0_{3\times 1}\\
        0_{1\times 3} & \sigma_n^2
    \end{bmatrix}E^\mathrm{T}\in\mathbb{R}^{6\times 6}.
\end{align*}


With the state transition equation in \eqref{eq_state_transition} and the measurement equation in \eqref{eq_final_measurement_eq}, the bearing-strength estimator can be obtained based on the pseudo-linear Kalman filter framework~\cite{Kulikov2018, Ning2023, Ning2024}. 


\section{Observability analysis}

It is important to theoretically derive the observability condition of the bearing-strength method to determine whether the additional bearing-related measurement can enhance the system's observability.

\subsection{Problem formulation}
The observability problem that we aim to solve is whether $p_e(t)$ can be recovered from $p_s(t)$, $g(t)$, and $n(t)$. We consider a general case where the motion of the emitter and the sensor can be described by $k$th-order polynomials:
\begin{align*}
    p_e(t) &= b_0 +b_1t +\cdots +b_kt^k,\\
    p_s(t) & = c_0 + c_1t +\cdots +c_kt^k +h(t),
\end{align*}
where, $\{b_i\}^k_{i=0}\in\mathbb{R}^3$ are unknown constant vectors, $\{c_i\}^k_{i=0}\in\mathbb{R}^3$ are known constant parameters, and
\begin{align}
    \label{eq_ht}
    h(t) = d_1t^{k+1} + d_2t^{k+2} + \cdots
\end{align}
represents \emph{higher-order} motion, where $\{d_i\}^\infty_{i=k}\in\mathbb{R}^3$ are constant vectors.
Let $s(t)\in\mathbb{R}^3$ be the relative motion between the target and the observer:
\begin{align}
    \label{eq_st}
    s(t)\dot{=}p_e(t)-p_s(t)=s_0+s_1t+\cdots+s_kt^k+h(t),
\end{align}
where $s_i=b_i-c_i\in\mathbb{R}^3$.

If we can determine the values of $\{s_i\}^k_{i=0}$, then $\{b_i\}^k_{i=0}$ and hence the emitter's motion can be determined.
It follows from~\eqref{eq_g_nonlinear_mear_eq} and~\eqref{eq_strength_nonlinear_mear_eq} that $p_e(t) - p_s(t) = g(t)\sqrt{n_e/n(t)}$.
Substituting it into~\eqref{eq_st} yields
\begin{align}
    \label{eq_linear_eqs}
    s_0+s_1t+\cdots+s_kt^k+h(t)=g(t)\sqrt{n_e/n(t)}.
\end{align}
Here, $s_i$ and $n$ are unknowns to be determined and $g(t)$, $n(t)$, $h(t)$ are known.
Equation~\eqref{eq_linear_eqs} can be reorganized into a matrix form:
\begin{align}
    \label{eq_Linear_eq_matrix}
    A(t)X=h(t),
\end{align}
where
\begin{align}
    A(t) &=\begin{bmatrix}
        I, tI, \cdots, t^kI, \rho(t)
    \end{bmatrix}\in\mathbb{R}^{3\times(3k+4)},
    \label{eq_A2}\\
    X &=\begin{bmatrix}
        s_0^\mathrm{T}, \cdots, s_k^\mathrm{T}, \sqrt{n_e}
    \end{bmatrix}^\mathrm{T}\in\mathbb{R}^{3k+4}.
\end{align}
Here, $\rho(t)=-g(t)/\sqrt{n(t)}\in\mathbb{R}^3$.
Therefore, the observability analysis problem becomes solving the linear equations~\eqref{eq_Linear_eq_matrix}.

\subsection{Observability condition}

Equation~\eqref{eq_Linear_eq_matrix} is under-determined. Continuously taking derivation on both sides of~~\eqref{eq_Linear_eq_matrix}, and combining them gives an over-determined system:
\begin{align}
    \label{eq_barA_linear_eq}
    \bar{A}(t)X=\bar{h}(t),
\end{align}
where
\begin{align*}
    \bar{A} =\begin{bmatrix}
        A(t) \\ A^{'}(t) \\ \vdots \\ A^{(K)}(t)
    \end{bmatrix},
    &&
    \bar{h}(t) = \begin{bmatrix}
        h(t) \\ h^{'}(t) \\ \vdots \\ h^{(K)}(t)
    \end{bmatrix}.
\end{align*}
Substituting~\eqref{eq_A2} into $\bar{A}(t)$ yields
\begin{align*}\bar{A}(t)\rightarrow
\left[\begin{array}{cccc:c}
    I & tI & \cdots & t^kI & \rho(t) \\
    0 & I & \cdots & kt^{k-1}I & \rho^{'}{t}\\
    \vdots & \vdots & \ddots & \vdots & \vdots \\
    0 & 0 & \cdots & k!I & \rho^{(k)}{t} \\
    \hdashline
    0 & 0 & \cdots & 0 & \rho^{(k+1)}{t} \\
    \vdots & \vdots & \ddots & \vdots & \vdots \\
    0 & 0 & \cdots & 0 & \rho^{(K)}{t}\end{array}\right].
 \end{align*}
As a result, $\bar{A}(t)$ has full column rank if and only if there exists $i\in\{k+1, \cdots, K\}$ such that 
\begin{align}
    \label{eq_condition1}
    \rho^{(i)}(t)\neq 0.
\end{align}
Since $\rho(t)=-g(t)\sqrt{1/n(t)}$ as shown in~\eqref{eq_A2} and $g(t)\sqrt{n_e/n(t)}=s_0+s_1t+\cdots+s_kt^k+h(t)$ as shown in~\eqref{eq_linear_eqs}, we can rewrite~\eqref{eq_condition1} to
\begin{align*}
    -\dfrac{1}{\sqrt{n_e}}(s_0+\cdots+s_kt^k+h(t))^{(i)}\neq 0_{3\times1}.
\end{align*}
Since $i\geq k+1$, the above equation is equivalent to
\begin{align}
    \label{eq_condition2}
    h^{(i)}(t)\neq 0.
\end{align}
According to the definition of $h(t)$ in~\eqref{eq_ht}, the obsevability condition is
\begin{align*}
    h(t)\neq 0.
\end{align*}

\begin{figure*}[!t]
    \centering
    \subfloat[Scenario 1: Circular motion around the target. The bearing-strength approach converges effectively.]{
        \includegraphics[width=\linewidth]{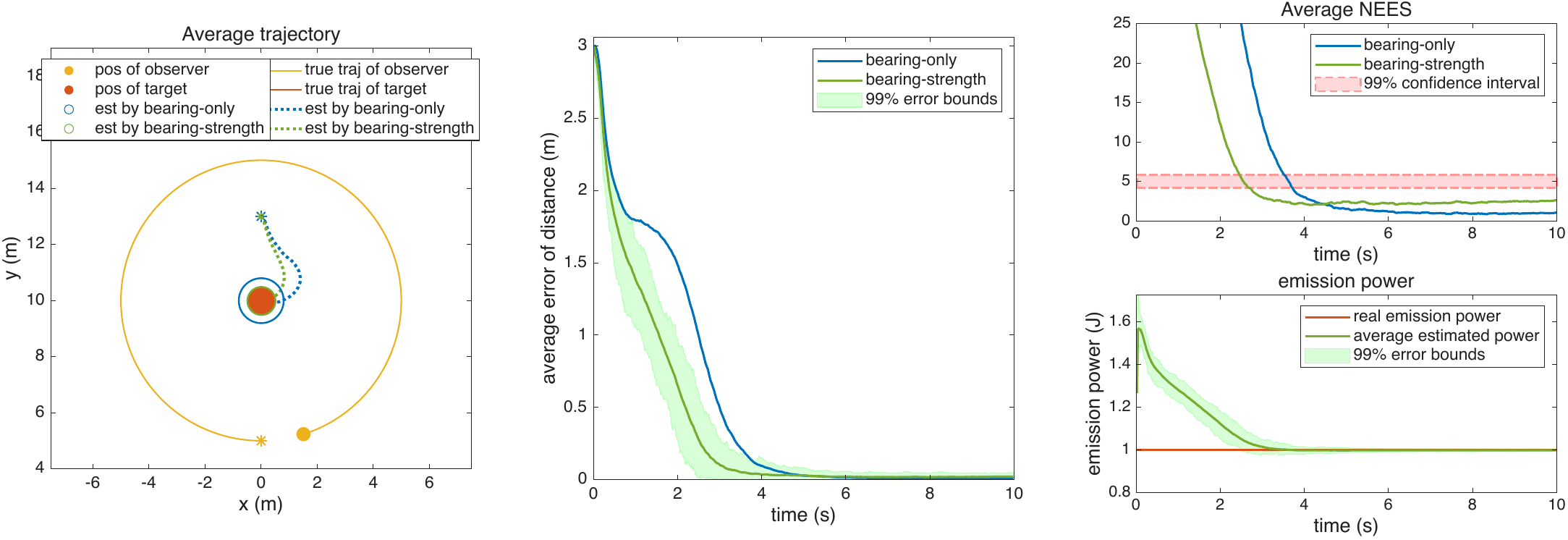}
        \label{fig:sim_exp1}
    }
    \hfill
    \subfloat[Scenario 2: Straight motion towards the target. The bearing-only approach fails, but the bearing-angle approach works effectively.]{
        \includegraphics[width=\linewidth]{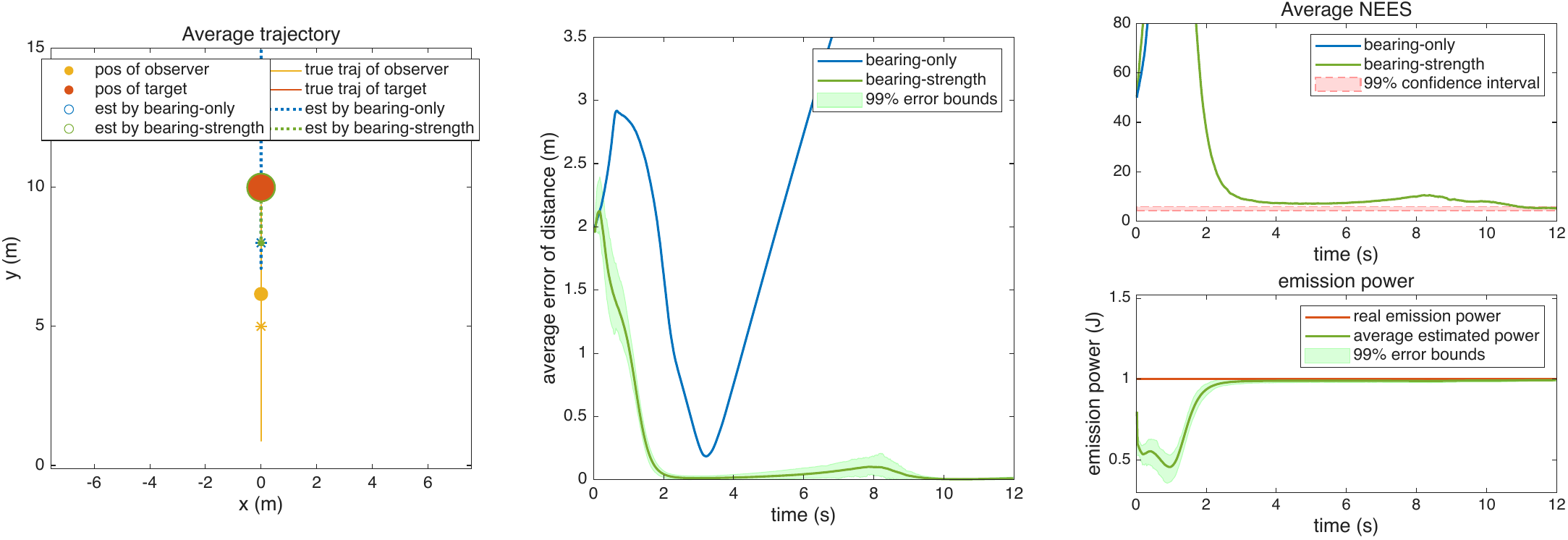}
        \label{fig:sim_exp2}
    }
    \caption{Numerical simulation results.}
    \label{fig:sim_results}
\end{figure*}

In summary, the observability condition of the bearing-strength method is that the sensor has a higher-order motion than the emitter. A key improvement over the traditional bearing-only method is that this higher-order motion is no longer required to be orthogonal to the bearing vector. This added flexibility allows the bearing-strength method to estimate the emitter's motion even in challenging scenarios, such as when the sensor moves directly towards the emitter.

Beyond the relaxed geometric requirements, this property also enhances the system's robustness against the physical characteristics of the emitter. For instance, although some emitters like LEDs exhibit anisotropic radiation patterns (i.e., $n = G(\theta, \phi) \cdot n_e / d^2$), the relative orientation $(\theta, \phi)$ remains constant during the aforementioned radial trajectories. In such cases, the directional gain $G(\theta, \phi)$ can be lumped into the unknown constant $n'_e$. Since our filter is designed to estimate the unknown power online, it naturally compensates for the directional gain in these geometrically degenerate scenarios, where the bearing-only method typically fails.

\section{Numerical Simulation Results}
\label{sec:simulation}

This section presents a set of numerical simulation results to demonstrate the effectiveness of the proposed bearing-strength approach. 

The values of the parameters in the two estimators are selected as $\sigma_v = 10^{-3}$, $\sigma_n = 0.01$, and $\sigma_\mu = 0.01$. The initial covariance matrix of the estimated states is set to $P(t_0) = 0.1I$ for both methods. The target is modeled as a stationary light source with a normalized true emission power of $n_e = 1$. The update rate of the system is $50\text{ Hz}$. We perform $N_x = 100$ Monte Carlo simulations for each scenario.

We use the normalized estimation error squared (NEES)\cite{Bar1998} to analyze the consistency of the estimation algorithms. In particular, the value of the average NEES is
\begin{equation}
\bar{\epsilon}_{\text{NEES}} = \frac{1}{N_x}\sum_{i=1}^{N_x} (x - \hat{x}_i)^T P_i^{-1} (x - \hat{x}_i),
\end{equation}
where $\hat{x}_i$ is the estimated state vector in the $i$-th simulation, and $P_i$ is the covariance matrix obtained from the estimator in the $i$-th simulation.

\subsection{Scenario 1: Circular motion around the target}
In the first scenario, the target is stationary and located at $p_T = [0, 10]^T$. The observer moves on a circle centered at the target with a speed of $3\text{ m/s}$ (see Fig.\ref{fig:sim_exp1}). The radius of the circle is $5\text{ m}$. The initial estimates of the observer's states are $\hat{p}_o(t_0) = [0, 13]^T$ and $\hat{v}_o(t_0) = [0, 0]^T$. The initial guess for the square root of the target's emission power is set to $\sqrt{\hat{n}_e(t_0)} = \sqrt{1.6}$. During this process, the bearing vector varies continuously due to the circular trajectory, while the received signal strength varies slightly around a mean value due to the measurement noise. This scenario is favorable to the conventional bearing-only approach because its fundamental observability condition — viewing the target from different angles — is well satisfied.

Fig.\ref{fig:sim_exp1} shows the estimation results obtained by the bearing-only and bearing-strength approaches. As can be seen, both algorithms perform well under this sufficient lateral excitation. The convergence of the bearing-strength approach is stable, and it successfully estimates the emission power of the target alongside its kinematic states.

\subsection{Scenario 2: Straight motion towards the target}
In the second scenario, the target is also stationary, but the observer moves along a straight line towards the target (see Fig.\ref{fig:sim_exp2}). During this process, the bearing vector remains strictly constant, while the received signal strength increases rapidly as the distance decreases. This scenario is widely known as the most challenging case for the bearing-only approach because its observability condition (i.e., the requirement for lateral motion) is completely unfulfilled.

In this simulation scenario, the observer moves along a straight line towards the target with a constant acceleration of $2\text{ m/s}^2$. The initial conditions are $v_o(t_0) = [0, 4]^T$ and $p_o(t_0) = [0, 5]^T$. The initial estimates are the same as those in Scenario 1. In this scenario, the true bearing of the target relative to the observer remains unchanged, causing a severe geometric singularity in traditional methods.

Fig.\ref{fig:sim_exp2} shows the estimation results of the bearing-only and bearing-strength approaches. As theoretically expected, the conventional bearing-only approach diverges rapidly since its fundamental observability condition is not satisfied. By contrast, the proposed bearing-strength approach converges effectively. It is able to localize the target and precisely estimate its emission power, demonstrating the strong observability enhancement of the bearing-strength approach even under pure radial motion.

\begin{figure*}[!t]
    \centering
    \subfloat[Linearity test: The grayscale camera shows a linear relationship between gray values and light strength, while the RGB camera exhibits non-linearity due to internal ISP.]{
        \includegraphics[width=0.48\linewidth]{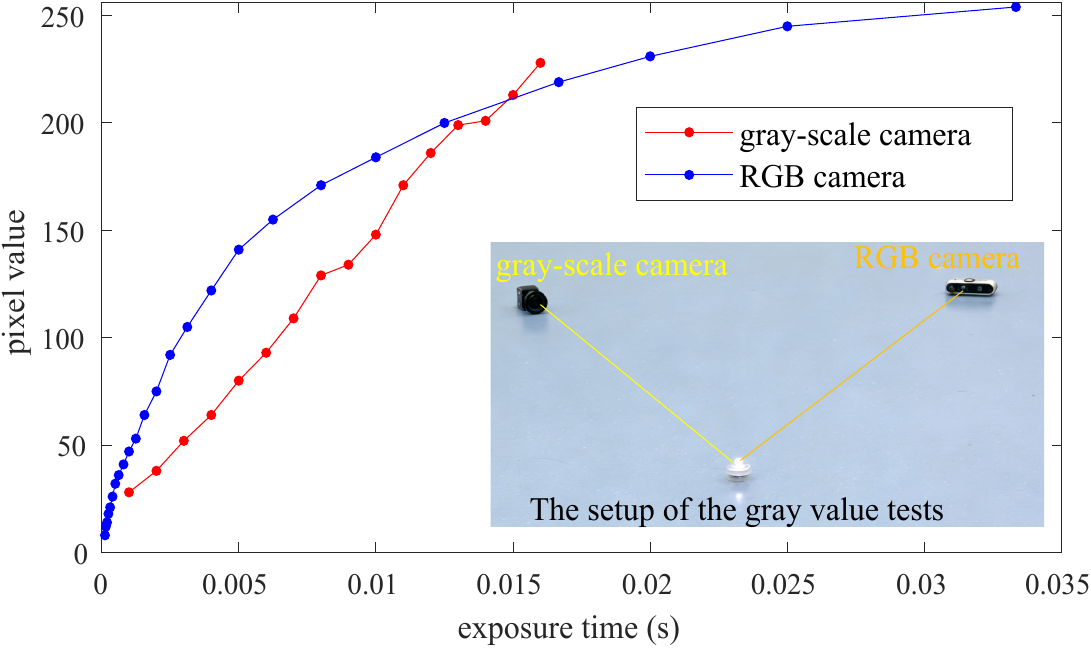}
        \label{fig_test_scenario}
    }
    \hfill
    \subfloat[Experimental system: The setup includes a point light source (emitter) on a mobile robot and a handheld grayscale camera (sensor) in a dynamic tracking task.]{
        \includegraphics[width=0.48\linewidth]{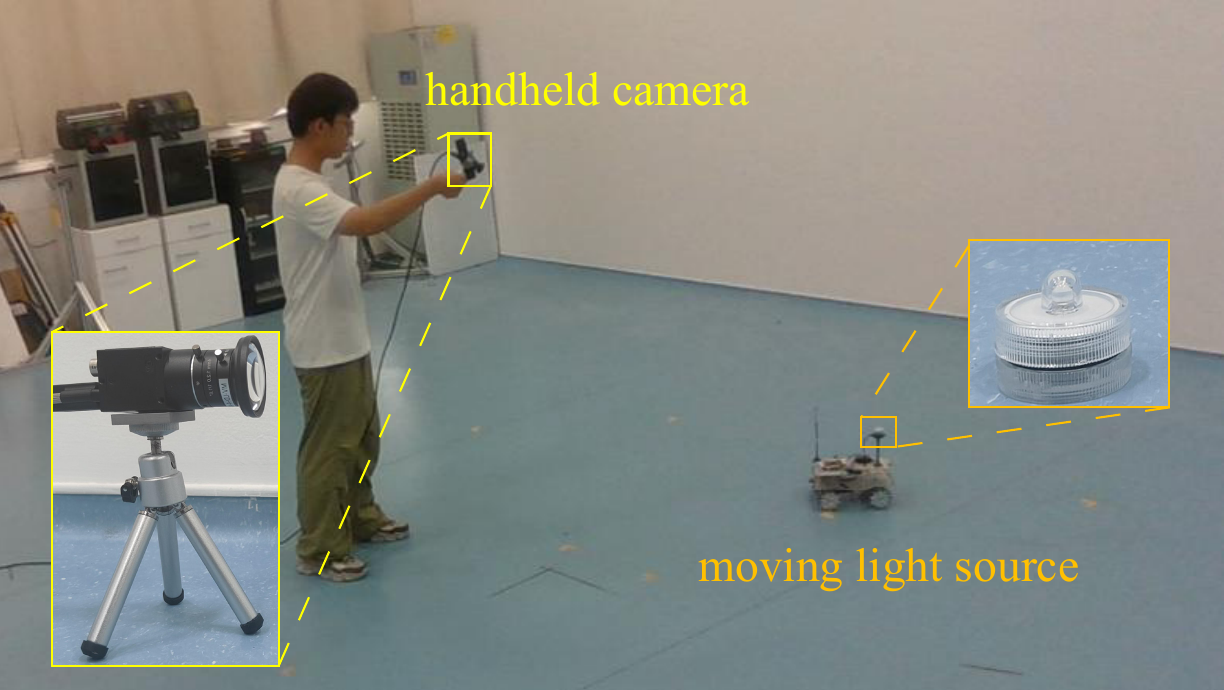}
        \label{fig_experiment_scenario}
    }
    \caption{Experimental validation of sensor characteristics and the real-world tracking setup.}
    \label{fig_experimental_all}
\end{figure*}

\section{Real-World Experimental Results}
This section presents two sets of real-world experimental results to verify the effectiveness of the proposed method.

\subsection{Experimental setup}

We intentionally selected light sources as the representative energy emitters for our real-world experiments. Light sources represent a more challenging case due to their inherent anisotropy and susceptibility to environmental light compared to other sources. If the bearing-strength method can effectively estimate the motion of such complex and non-ideal emitters, its performance on more isotropic sources, like acoustic or omnidirectional radio signals, can be reasonably expected to be even more robust, as the constant-power assumption would be more naturally satisfied across various observation angles.

The camera and the light source used in the experiments are shown in Fig.~\ref{fig_test_scenario}.
The sensor is a grayscale camera with a resolution of 640$\times$480 pixels.
The lens provides a 90-degree field of view with little distortion.
The camera's intrinsic parameters are calibrated beforehand.
The camera's exposure parameters, including ISO, aperture, and exposure time, are manually set in advance and remain fixed during the experiments, ensuring that the gray values are influenced only by received light strength.
The Vicon motion capture system is used to provide the pose of the camera. This can be replaced by other positioning systems in practice.
The position of the light source provided by Vicon is treated as the ground-truth value for calculating measurement noises (see the right subfigures of Fig.~\ref{fig_exp}).

The experimental setup is shown in Fig.~\ref{fig_experiment_scenario}.
The camera is handheld by a human operator.
The light source is mounted on a ground robot.
The robot is manually controlled via a remote controller.
Its speed ranges from 0.1 to 0.7 m/s, with an average speed of 0.2 m/s.

\subsection{Image gray value vs. received light strength }
\label{sec_cam_test}

Before conducting the experiments, we performed a preliminary test to verify the linear relationship between gray values and the received light strength.
Both grayscale and RGB cameras are tested for comparison.
The experimental setup is shown in Fig.~\ref{fig_test_scenario}.
The cameras are positioned in front of the light source, and their exposure times are adjusted to simulate a linear increase in the received light strength.

Fig.~\ref{fig_experiment_scenario} shows the curves that represent the relationship between gray values and exposure times.
As can be seen, the gray value of the grayscale camera is proportional to the received strength, whereas the RGB camera does not follow this rule.
This is because the ISP processing in RGB cameras, including white balance and gamma correction, alters the original gray values. These factors make the inverse square law invalid for the RGB camera.
As a result, we use the grayscale camera in the following experiments.

\subsection{Experimental results}
\begin{figure*}[!t]
    \centering
    \subfloat[Experiment 1: The sensor moves around the emitter. Both the bearing-only and bearing-strength methods work well.]{
        \includegraphics[width=\linewidth]{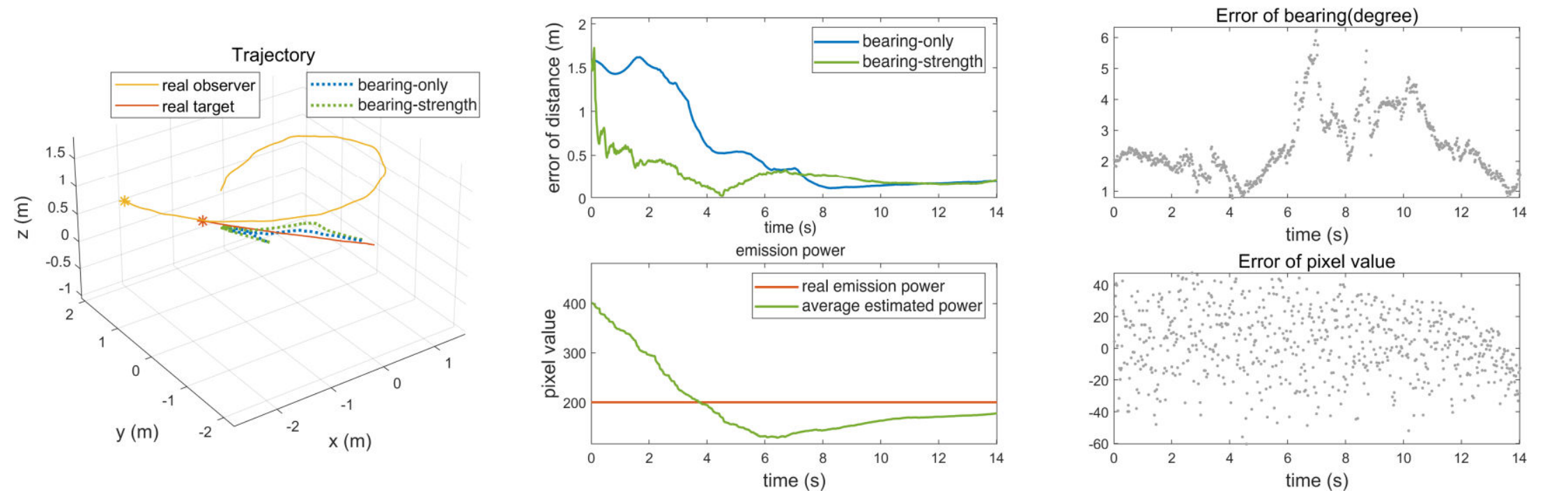}
        \label{fig_exp_1}
    }
    \hfill
    \subfloat[Experiment 2: The sensor moves close and far from the light source periodically. The bearing-strength method performs effectively, but the bearing-only method works unstably.]{
        \includegraphics[width=\linewidth]{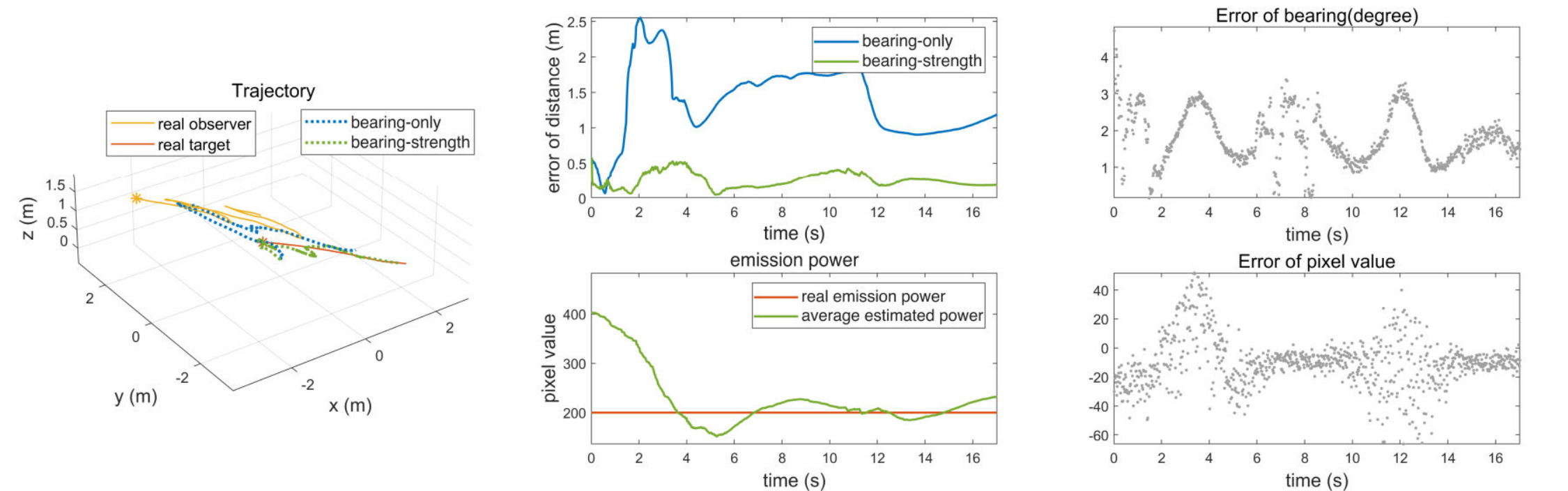}
        \label{fig_exp_2}
    }
    \caption{Two experimental results.}
    \label{fig_exp}
\end{figure*}

Two sets of experiments have been conducted.
Both bearing-only and bearing-strength estimators are implemented for comparison.
We use the same parameter values in both estimators and across two experiments.
The variances are selected as $\sigma_a = 0.01$, $\sigma_\text{strength}=10^{-4}$, $\sigma_g = 0.2$, and $\sigma_n = 2$.
The initial covariance matrix of the estimated states is set to $P(t_0)=4I$.
The update rate of the estimator is 50~Hz.
Regarding computational efficiency, the proposed bearing-strength method exhibits a sub-millisecond latency, offering a significant advantage over both conventional vision-based bearing-only and bearing-angle approaches \cite{Ning2024}. While those baseline methods typically necessitate computationally heavy image recognition to identify the full target or extract complex geometric features, our approach simply tracks the brightest optical flare. This effectively eliminates the overhead of complex visual processing, making it highly efficient for resource-constrained platforms.
Better performances can be achieved if the parameters are well-tuned for specific scenarios.
This experiment is conducted in a low-light environment, allowing the light source to be detected by identifying the brightest point in the image.


The experimental results are shown in Fig.~\ref{fig_exp}.

In the first experiment, the camera is held about 1.5~m above the ground and moves around the emitter. 
The distance between the sensor and the light source varies from 1.7 to 2.6~m.
In this experiment, the bearing vector varies sufficiently, and hence the observability conditions for the bearing-only and bearing-strength methods are both well satisfied.
As shown in Fig.~\ref{fig_exp_1}, both methods perform well, and the bearing-strength method performs slightly better than the bearing-only one.

In the second experiment, the camera moves approximately along the bearing vector close to and far from the light source periodically.

To quantitatively evaluate the stability and accuracy of the proposed bearing-strength method, the Root Mean Square Errors (RMSE) of the measurement residuals for both real-world experiments are summarized in Table \ref{tab:real_rmse}. 

\begin{table}[htbp]
\caption{Measurement Residual RMSE of the Bearing-Strength Method in Real-World Experiments}
\label{tab:real_rmse}
\begin{center}
\begin{tabular}{lcc}
\hline
Scenario & Bearing Error ($^\circ$) & Pixel Value Error \\
\hline
Experiment 1 (Circular Motion) & 2.02 & 23.2 \\
Experiment 2 (Radial Motion)   & 2.51 & 24.3 \\
\hline
\end{tabular}
\end{center}
\end{table}

In this experiment, the strength measurement $\tilde{n}$ varies significantly, while the bearing does not.
Without surprise, the bearing-only method performs unstably due to weak observability (Fig.~\ref{fig_exp_2}).
By contrast, the bearing-strength method performs stably due to its enhanced observability.

\section{Discussion}
While the proposed framework demonstrates strong performance, practical implementations must account for certain nonidealities. 

First, regarding the RSS-distance mismatch, commercial cameras often introduce nonlinearities due to internal ISP algorithms. As validated in Section V-B, we mitigate this by employing a grayscale camera with fixed exposure. Other environmental nonidealities, such as light scattering or target anisotropy, could be addressed in future work by incorporating an environmental scattering coefficient into the measurement equations.

Second, regarding the unknown transmit power $n_e$, if the emitter experiences slow power fluctuations like battery degradation, the estimator naturally tracks these changes as $\sqrt{n_e}$ is explicitly estimated online. Rapid, high-frequency power variations would manifest as additional measurement noise, which the filter accommodates through the tuning of the noise covariance matrix $\Sigma_\nu$.

\section{Conclusion}

Motivated by the limitations of the traditional bearing-only method, this paper proposed a novel bearing-strength method to estimate the motion of the moving energy emitter.
We showed that the bearing-strength method significantly enhanced the system's observability compared to the bearing-only method.
The sensor's extra lateral motion can be significantly relaxed.
As shown in real-world experiments, the bearing-strength method can successfully estimate the emitter's motion while the bearing-only method fails.
The enhanced observability of the bearing-strength method comes with no additional costs since the received strength measurement inherently exists in passive sensors.
Moreover, the proposed bearing-strength estimator is a unified method that can be applied to light, acoustic, and radio emitters.

\bibliography{ref}

@Article{Ning2024a,
  author    = {Ning, Zian and Zhang, Yin and Lin, Xiaofeng and Zhao, Shiyu},
  journal   = {Unmanned Systems},
  title     = {A Real-to-Sim-to-Real Approach for Vision-Based Autonomous MAV-Catching-MAV},
  year      = {2024},
  issn      = {2301-3869},
  month     = may,
  number    = {04},
  pages     = {787--798},
  volume    = {12},
  doi       = {10.1142/s2301385025500360},
  publisher = {World Scientific Pub Co Pte Ltd},
}

@Article{Ning2024,
  author    = {Zian Ning and Yin Zhang and Jianan Li and Zhang Chen and Shiyu Zhao},
  journal   = {The International Journal of Robotics Research},
  title     = {A bearing-angle approach for unknown target motion analysis based on visual measurements},
  year      = {2024},
  issn      = {1741-3176},
  month     = jul,
  number    = {8},
  pages     = {1228--1249},
  volume    = {43},
  doi       = {10.1177/02783649241229172},
  publisher = {SAGE Publications},
}

@Article{Ning2023,
  author    = {Zian Ning and Yin Zhang and Shiyu Zhao},
  journal   = {Control Theory and Technology},
  title     = {Comparison of different pseudo-linear estimators for vision-based target motion estimation},
  year      = {2023},
  month     = aug,
  number    = {3},
  pages     = {448--457},
  volume    = {21},
  doi       = {10.1007/s11768-023-00161-y},
  publisher = {Springer Science and Business Media {LLC}},
}

@Article{Li2023,
  author    = {Jianan Li and Zian Ning and Shaoming He and Chang-Hun Lee and Shiyu Zhao},
  journal   = {{IEEE} Transactions on Robotics},
  title     = {Three-Dimensional Bearing-Only Target Following via Observability-Enhanced Helical Guidance},
  year      = {2023},
  issn      = {1941-0468},
  month     = apr,
  number    = {2},
  pages     = {1509--1526},
  volume    = {39},
  doi       = {10.1109/tro.2022.3218268},
  publisher = {Institute of Electrical and Electronics Engineers ({IEEE})},
}

@Article{Qiu2019,
  author    = {Kejie Qiu and Tong Qin and Wenliang Gao and Shaojie Shen},
  journal   = {{IEEE} Transactions on Robotics},
  title     = {Tracking 3-{D} Motion of Dynamic Objects Using Monocular Visual-Inertial Sensing},
  year      = {2019},
  month     = aug,
  number    = {4},
  pages     = {799--816},
  volume    = {35},
  doi       = {10.1109/tro.2019.2909085},
  publisher = {Institute of Electrical and Electronics Engineers ({IEEE})},
}

@InProceedings{Dressel2019,
  author    = {Louis Dressel and Mykel J. Kochenderfer},
  booktitle = {2019 International Conference on Robotics and Automation ({ICRA})},
  title     = {Hunting drones with other drones: tracking a moving radio target},
  year      = {2019},
  month     = may,
  comment   = {无人机反制，感知：无线电信号，估计：Particle Filter，跟踪控制：马尔可夫决策过程，因为目标的估计是不准确的，亮点：使用马尔可夫过程来做抓捕无人机的行为决策},
  doi       = {10.1109/icra.2019.8794243},
}

@InProceedings{wuller2007,
  author       = {Dietmar W{\"u}ller and Helke Gabele},
  booktitle    = {Digital Photography III},
  title        = {The usage of digital cameras as luminance meters},
  year         = {2007},
  editor       = {Russel A. Martin and Jeffrey M. DiCarlo and Nitin Sampat},
  month        = feb,
  organization = {International Society for Optics and Photonics},
  pages        = {65020U},
  publisher    = {SPIE},
  volume       = {6502},
  doi          = {10.1117/12.703205},
}

@Article{hiscocks2011measuring,
  author  = {Hiscocks, Peter D and Eng, P},
  journal = {Syscomp Electronic Design Limited},
  title   = {Measuring luminance with a digital camera},
  year    = {2011},
  volume  = {2},
}

@Article{Hoelzer1978,
  author  = {Hoelzer, HD and Johnson, GW and Cohen, AO},
  journal = {IR \& D Report},
  title   = {Modified polar coordinates-the key to well behaved bearings only ranging},
  year    = {1978},
  pages   = {78--M19},
}

@Article{Voudoukis_Oikonomidis_2017,
  author = {Nikolaos Voudoukis and Sarantos Oikonomidis},
  title  = {Inverse Square Law for Light and Radiation: A Unifying Educational Approach},
  year   = {2017},
  month  = nov,
  pages  = {23–27},
  volume = {2},
  doi    = {10.24018/ejeng.2017.2.11.517},
  url    = {https://ej-eng.org/index.php/ejeng/article/view/517},
}

@Article{Zhao2019,
  author    = {Shiyu Zhao and Daniel Zelazo},
  journal   = {{IEEE} Control Systems Magazine},
  title     = {Bearing Rigidity Theory and Its Applications for Control and Estimation of Network Systems: Life Beyond Distance Rigidity},
  year      = {2019},
  month     = apr,
  number    = {2},
  pages     = {66--83},
  volume    = {39},
  doi       = {10.1109/mcs.2018.2888681},
  publisher = {Institute of Electrical and Electronics Engineers ({IEEE})},
}

@Article{Kulikov2018,
  author    = {Gennady Yu. Kulikov and Maria V. Kulikova},
  journal   = {IET Control Theory \& Applications},
  title     = {Moore-Penrose-pseudo-inverse-based Kalman-like filtering methods for estimation of stiff continuous-discrete stochastic systems with ill-conditioned measurements},
  year      = {2018},
  month     = sep,
  number    = {16},
  pages     = {2205--2212},
  volume    = {12},
  doi       = {10.1049/iet-cta.2018.5404},
  publisher = {Institution of Engineering and Technology ({IET})},
}

@Article{Fogel1988,
  author    = {E. Fogel and M. Gavish},
  journal   = {{IEEE} Transactions on Aerospace and Electronic Systems},
  title     = {Nth-order dynamics target observability from angle measurements},
  year      = {1988},
  month     = may,
  number    = {3},
  pages     = {305--308},
  volume    = {24},
  doi       = {10.1109/7.192098},
  publisher = {Institute of Electrical and Electronics Engineers ({IEEE})},
}

@Article{He2019,
  author    = {Shaoming He and Hyo-Sang Shin and Antonios Tsourdos},
  journal   = {{IEEE} Transactions on Robotics},
  title     = {Trajectory Optimization for Target Localization With Bearing-Only Measurement},
  year      = {2019},
  month     = jun,
  number    = {3},
  pages     = {653--668},
  volume    = {35},
  doi       = {10.1109/tro.2019.2896436},
  publisher = {Institute of Electrical and Electronics Engineers ({IEEE})},
}

@article{zanella2016best,
  title={Best practice in {RSS} measurements and ranging},
  author={Zanella, Andrea},
  journal={IEEE Communications Surveys \& Tutorials},
  volume={18},
  number={4},
  pages={2662--2686},
  year={2016},
  publisher={IEEE}
}

@article{wang2011new,
  title={A new approach to sensor node localization using {RSS} measurements in wireless sensor networks},
  author={Wang, Gang and Yang, Kehu},
  journal={IEEE Transactions on Wireless Communications},
  volume={10},
  number={5},
  pages={1389--1395},
  year={2011},
  publisher={IEEE}
}

@article{tomic20163,
  title={{3-D} target localization in wireless sensor networks using {RSS} and {AoA} measurements},
  author={Tomic, Slavisa and Beko, Marko and Dinis, Rui},
  journal={IEEE Transactions on Vehicular Technology},
  volume={66},
  number={4},
  pages={3197--3210},
  year={2016},
  publisher={IEEE}
}

@article{guo2024global,
  title={Global-Local MAV Detection Under Challenging Conditions Based on Appearance and Motion},
  author={Guo, Hanqing and Zheng, Ye and Zhang, Yin and Gao, Zhi and Zhao, Shiyu},
  journal={IEEE Transactions on Intelligent Transportation Systems},
  year={2024},
  publisher={IEEE}
}

@article{zhang2024domain,
  title={Domain Adaptive Detection of MAVs: A Benchmark and Noise Suppression Network},
  author={Zhang, Yin and Deng, Jinhong and Liu, Peidong and Li, Wen and Zhao, Shiyu},
  journal={IEEE Transactions on Automation Science and Engineering},
  year={2024},
  publisher={IEEE}
}

@ARTICLE{10659110,
  author={Pliska, Michal and Vrba, Matouš and Báča, Tomáš and Saska, Martin},
  journal={IEEE Robotics and Automation Letters}, 
  title={Towards Safe Mid-Air Drone Interception: Strategies for Tracking \& Capture}, 
  year={2024},
  volume={9},
  number={10},
  pages={8810-8817},
  doi={10.1109/LRA.2024.3451768}
}

@Article{Shi2018,
  author    = {Xiufang Shi and Chaoqun Yang and Weige Xie and Chao Liang and Zhiguo Shi and Jiming Chen},
  journal   = {{IEEE} Communications Magazine},
  title     = {Anti-drone system with multiple surveillance technologies: architecture, implementation, and challenges},
  year      = {2018},
  month     = apr,
  number    = {4},
  pages     = {68--74},
  volume    = {56},
  comment   = {团队：浙大老师， 地面固定设备，感知：声音、无线电、视觉，估计：DOA、RSS算法，反制：地面 - 干扰。无人机反制，反无人机，},
  doi       = {10.1109/mcom.2018.1700430},
  publisher = {Institute of Electrical and Electronics Engineers ({IEEE})},
}

@Article{Vrba2020,
  author    = {Matous Vrba and Martin Saska},
  journal   = {{IEEE} Robotics and Automation Letters},
  title     = {Marker-less micro aerial vehicle detection and localization using convolutional neural networks},
  year      = {2020},
  month     = apr,
  number    = {2},
  pages     = {2459--2466},
  volume    = {5},
  doi       = {10.1109/lra.2020.2972819},
  publisher = {Institute of Electrical and Electronics Engineers ({IEEE})},
}

@ARTICLE{Bar1998,
  author={Bar-Shalom, Y. and Xiao-Rong Li},
  journal={IEEE Antennas and Propagation Magazine}, 
  title={Estimation and Tracking: Principles, Techniques, and Software}, 
  year={1996},
  volume={38},
  number={1},
  pages={62-},
  doi={10.1109/MAP.1996.491294}}
\bibliographystyle{ieeetr}
\end{document}